\pdfoutput=1
\documentclass[letterpaper, 10 pt, conference]{ieeeconf}
\IEEEoverridecommandlockouts
\usepackage{lipsum}

\usepackage{multibib}
\newcites{app}{Appendix References}

\usepackage{wrapfig}
\usepackage{xcolor} 
\usepackage{stfloats}

\usepackage{amsmath}
\usepackage{amsfonts}
\usepackage{amsthm}
\usepackage{amssymb}
\usepackage{mathrsfs}  
\usepackage{bm}
\usepackage[ruled,vlined]{algorithm2e}
\usepackage{mathdots} 
\usepackage{dsfont} 
\usepackage{accents} 
\usepackage{tikz-cd} 
\usepackage{adjustbox} 
\usepackage{mathtools} 
\usepackage{cite} 

\let\labelindent\relax
\usepackage{enumitem}

\usepackage{caption}
\usepackage{multirow}
\usepackage{multicol}
\usepackage{mdframed} 

\usepackage{listings}
\usepackage{siunitx}

\usepackage{graphicx}

\usepackage{balance}

\newcommand{\comment}[1]{} 
 
\newcommand{\norm}[1]{\left\lVert#1\right\rVert} 
\newcommand\eqs[1]{\begin{equation}\begin{split}#1\end{split}\end{equation}} 
\newcommand\eqsnn[1]{\begin{equation*}\begin{split}#1\end{split}\end{equation*}} 
\newcommand\eqa[1]{\begin{align}#1\end{align}} 

\DeclareMathOperator*{\diag}{diag} 
\DeclareMathOperator*{\argmin}{arg\,min}

\newcommand\braces[1]{\left\{#1\right\}} 
\newcommand\brackets[1]{\left[#1\right]} 
\newcommand\parens[1]{\left(#1\right)} 

\newcommand\R{\mathbb{R}} 
\newcommand\N{\mathbb{N}} 

\newtheorem*{definition*}{Definition}

\usepackage[bookmarks=true, hidelinks]{hyperref}

\makeatletter
\let\save@mathaccent\mathaccent
\newcommand*\if@single[3]{%
  \setbox0\hbox{${\mathaccent"0362{#1}}^H$}%
  \setbox2\hbox{${\mathaccent"0362{\kern0pt#1}}^H$}%
  \ifdim\ht0=\ht2 #3\else #2\fi
  }
\newcommand*\rel@kern[1]{\kern#1\dimexpr\macc@kerna}
\newcommand*\widebar[1]{\@ifnextchar^{{\wide@bar{#1}{0}}}{\wide@bar{#1}{1}}}
\newcommand*\wide@bar[2]{\if@single{#1}{\wide@bar@{#1}{#2}{1}}{\wide@bar@{#1}{#2}{2}}}
\newcommand*\wide@bar@[3]{%
  \begingroup
  \def\mathaccent##1##2{%
    \let\mathaccent\save@mathaccent
    \if#32 \let\macc@nucleus\first@char \fi
    \setbox\z@\hbox{$\macc@style{\macc@nucleus}_{}$}%
    \setbox\tw@\hbox{$\macc@style{\macc@nucleus}{}_{}$}%
    \dimen@\wd\tw@
    \advance\dimen@-\wd\z@
    \divide\dimen@ 3
    \@tempdima\wd\tw@
    \advance\@tempdima-\scriptspace
    \divide\@tempdima 10
    \advance\dimen@-\@tempdima
    \ifdim\dimen@>\z@ \dimen@0pt\fi
    \rel@kern{0.6}\kern-\dimen@
    \if#31
      \overline{\rel@kern{-0.6}\kern\dimen@\macc@nucleus\rel@kern{0.4}\kern\dimen@}%
      \advance\dimen@0.4\dimexpr\macc@kerna
      \let\final@kern#2%
      \ifdim\dimen@<\z@ \let\final@kern1\fi
      \if\final@kern1 \kern-\dimen@\fi
    \else
      \overline{\rel@kern{-0.6}\kern\dimen@#1}%
    \fi
  }%
  \macc@depth\@ne
  \let\math@bgroup\@empty \let\math@egroup\macc@set@skewchar
  \mathsurround\z@ \frozen@everymath{\mathgroup\macc@group\relax}%
  \macc@set@skewchar\relax
  \let\mathaccentV\macc@nested@a
  \if#31
    \macc@nested@a\relax111{#1}%
  \else
    \def\gobble@till@marker##1\endmarker{}%
    \futurelet\first@char\gobble@till@marker#1\endmarker
    \ifcat\noexpand\first@char A\else
      \def\first@char{}%
    \fi
    \macc@nested@a\relax111{\first@char}%
  \fi
  \endgroup
}
\makeatother

\begin{document}

\title{
    DROP: \textbf{\underline{D}}exterous \textbf{\underline{R}}eorientation via \textbf{\underline{O}}nline \textbf{\underline{P}}lanning
}

\author{
    Albert H. Li$^\dagger$, Preston Culbertson$^\ddagger$, Vince Kurtz$^\ddagger$, Aaron D. Ames$^{\dagger,\ddagger}$%
    \thanks{$\dagger$ A. H. Li and A. D. Ames are with the Department of Computing and Mathematical Sciences, California Institute of Technology, Pasadena, CA 91125, USA, \texttt{\{alberthli, ames\}@caltech.edu}.}%
    \thanks{$\ddagger$ P. Culbertson, V. Kurtz, and A. D. Ames are with the Department of Civil and Mechanical Engineering, California Institute of Technology, Pasadena, CA 91125, USA, \texttt{\{pculbert, vkurtz, ames\}@caltech.edu}.}%
    \thanks{* This research is supported by Dow.}
}

\maketitle

\begin{abstract}

Achieving human-like dexterity is a longstanding challenge in robotics, in part due to the complexity of planning and control for contact-rich systems. In reinforcement learning (RL), one popular approach has been to use massively-parallelized, domain-randomized simulations to learn a policy \textit{offline} over a vast array of contact conditions, allowing robust sim-to-real transfer. Inspired by recent advances in real-time parallel simulation, this work considers instead the viability of \textit{online planning} methods for contact-rich manipulation by studying the well-known in-hand cube reorientation task. We propose a simple architecture that employs a sampling-based predictive controller and vision-based pose estimator to search for contact-rich control actions online. We conduct thorough experiments to assess the real-world performance of our method, architectural design choices, and key factors for robustness, demonstrating that our simple sampling-based approach achieves performance comparable to prior RL-based works. Supplemental material: \href{https://caltech-amber.github.io/drop}{https://caltech-amber.github.io/drop}.

\end{abstract}

\section{Introduction}\label{sec:intro}
Achieving dexterity comparable to human hands has been a longstanding challenge in robotics. While even simple robots can produce dynamic, contact-rich behavior \cite{ishihara_dynamic2006, chavan_extrinsic2014}, general methods for doing so are still scarce. For contact-rich tasks, reinforcement learning (RL) has been the dominant paradigm due to its ability to generate real-world robust plans. One well-studied task is in-hand cube reorientation, where a hand must rotate a cube to match consecutive goal orientations. Pioneered by OpenAI \cite{openai2018_dactyl} and extended by others \cite{handa2024_dextreme, chen2023_reorientation}, RL policies trained with massively-parallelized, domain-randomized simulations have achieved remarkably robust sim-to-real transfer for cube rotation. But, this \textit{offline} simulation-based approach requires substantial pre-execution computation and is inflexible to changing task specifications.

\begin{figure}[t]
\centering
\includegraphics[width=0.95\linewidth]{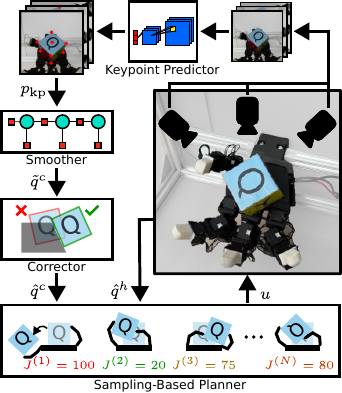}
\caption{
    \textbf{The DROP architecture.} DROP consists of (i) a vision-based cube pose estimator (composed of the Keypoint Predictor, Smoother, and Corrector), and (ii) a sampling-based planner that selects control actions by conducting model-based rollouts and iteratively improving the sampling distribution online based on the costs $J^{(i)}$.
}
\vspace{-0.5cm}
\label{fig:hero}
\end{figure}

Leveraging recent advances in real-time parallel simulation (e.g., MJPC) \cite{howell2022_mjpc}, we instead study the \textit{online} approach of sampling-based predictive control (SPC) methods for in-hand manipulation, which continuously replan by simulating parallel rollouts and applying optimal control actions over short time horizons. In contrast with RL, such online planning methods can adjust the task or model without re-training, but may demand expensive online computation. While tools like MJPC have made SPC feasible for many simulated contact-rich tasks, their real-world utility remains largely unproven.

Our main contribution is DROP (\textbf{\underline{D}}exterous \textbf{\underline{R}}eorientation via \textbf{\underline{O}}nline \textbf{\underline{P}}lanning), a system architecture (Fig. \ref{fig:hero}) that consists of (i) a simple sampling-based planner and (ii) a vision-based state estimator comprising a keypoint detection model, a pose smoother, and a collision-aware state corrector. Our aim is not to design the best cube reorientation policy, but to test the viability of sampling-based strategies by thoroughly assessing DROP's real-world performance, architectural design, and key factors for robustness. Still, DROP is the  first online planning method for cube reorientation on hardware, and we find that its performance is comparable with prior RL-based methods. For reproducibility, we provide open-source code, estimator weights, and hardware setup\footnote{Project website: \href{https://caltech-amber.github.io/drop}{https://caltech-amber.github.io/drop} \cite{li2024_drop_website}.}.

\section{Background and Preliminaries}

\subsection{Related Work}

The first viable methods for in-hand cube reorientation employed \textbf{\emph{reinforcement learning (RL)}}, starting with Dactyl \cite{openai2018_dactyl}, which popularized domain randomization for sim-to-real transfer of RL policies. Subsequent research built on Dactyl with improved domain randomization \cite{handa2024_dextreme}, tactile-only sensing \cite{yin2023_rotating, rostel2023estimator}, and ``palm-down'' rotations for various objects \cite{chen2021_generalreorientation, chen2023_reorientation}. While typically limited to specific robots or object classes, RL's robust hardware performance has established it as the standard for in-hand manipulation.

Others have explored model-based approaches to contact-rich manipulation via \textbf{\emph{contact-implicit trajectory optimization}} (CITO). Originally used in locomotion \cite{posa2014_direct_cito, doshi_cito_locomotion}, recent work \cite{lecleach2023_cimpc, howell2022_mjpc} shows that such planners can generate motions for simple contact-rich tasks like bimanual lifting \cite{kurtz2023_idto} or real-time planar rolling and sliding \cite{aydinoglu2022_c3, yang2024_dynamic}. Other approaches have combined CITO with learned signed distance fields \cite{yang2024_contactsdf}, global search \cite{chen2021_trajectotree}, or smoothed dynamics \cite{pang2023_quasistatic}.

\textbf{\emph{Sampling-based planning}} has a long history in manipulation, starting with methods like RRT, PRM, and STOMP \cite{lavalle2000_rrt, kavraki1996_prm, kalakrishnan2011_stomp}. Recent advances in SPC \cite{williams2018_mppi, jankowski2023_vpsto} have enabled simple contact-rich manipulation like pushing, sliding, and pick and place \cite{pezzato2023_isaac_mppi, bhardwaj2021_storm}, while new tools for real-time parallel simulation of contact-rich tasks \cite{howell2022_mjpc}, have made SPC viable for simulated contact-rich tasks with remarkably simple controllers. Still, little evidence exists for the viability of SPC for real-world contact-rich manipulation (concurrently, some works have studied it for locomotion \cite{alvarezpadilla2024realtimewholebodycontrollegged}), and overall, model-based methods (like SPC or CITO) have seldom studied tasks as hard as cube reorientation, which exhibits many simultaneous unpredictable contact modes.

Lastly, high-quality \textbf{\emph{object tracking}} is key for most of these methods, especially model-based online search. RL approaches \cite{openai2018_dactyl, handa2024_dextreme} usually use single-object pose estimation networks, but recent computer vision advances suggest that more general tracking pipelines like frame-to-frame point trackers \cite{doersch2024_bootstap, karaev2023_cotracker, xiao2024_spatialtracker} and foundation models for rigid body pose prediction \cite{wen2024_foundationpose} may boost performance.

\subsection{Mathematical Notation}

We model the continuous-time cube-in-hand dynamics as
\begin{equation}\label{eq:dynamics}
    \dot{x}(t) = f\left( x(t), u(t) \right),
\end{equation}
where $x$ is the state and $u$ are control inputs. We do not assume access to a closed-form expression for $f$, but instead to a simulator (MuJoCo \cite{todorov2012mujoco}) that generates state trajectories $x(t)$ given an initial state $x_0$ and a control sequence $u(t)$.

The state $x$ consists of the generalized positions and velocities of both the cube ($q^c, v^c$) and hand ($q^h, v^h$):
\begin{equation}\label{eq:state}
    x = [q, v] = [q^c, q^h, v^c, v^h].
\end{equation}
In this work, the control actions $u$ are position setpoints for the hand, which are tracked by a lower-level PD controller internal to the model $f$. Following \cite{howell2022_mjpc}, we represent a control trajectory $u(t)$ on $t \in [0, T]$ with $K$ spline knots,
\eqa{
    U &= [u_1, u_2, \dots, u_k, \dots, u_{K}],
}
which constitute the decision variables for online planning. The control objective is encoded by a cost functional
\begin{equation}\label{eqn:generic_cost}
    J(U; x_0) = \int_0^T \ell\left(x(t), u(t)\right)dt,
\end{equation}
where querying $J(U; x_0)$ requires simulating $f$ using an open-loop control signal $u(t)$ from initial condition $x_0$.

Throughout the paper, we use the following notation to denote ``subtraction'' of two quaternions $a,b\in\mathbb{H}$:
\eqs{
    a \ominus b := \text{Log}\parens{a \circ b^{-1}} \in \mathfrak{so}(3),
}
and similarly, when $a,b\in SE(3)$, $a \ominus b \in \mathfrak{se}(3)$.

\section{The DROP Algorithm}\label{sec:method}

DROP consists of (i) a sampling-based planner and (ii) a keypoint-based cube pose estimator (see Fig.~\ref{fig:hero}). The planner continually updates the control spline $U$ via SPC. We use MJPC \cite{howell2022_mjpc} to handle both parallel rollouts and policy updates.

\subsection{Sampling-Based Planning and Control}\label{sec:control}

Algorithm~\ref{alg:sampling_planner} describes a generic SPC procedure. At time $t \in \mathbb{R}$, the planner receives a state estimate $\hat{x}(t)$ and rolls out a batch of $N$ open-loop control sequences $U^{(i)}$ drawn from some parametric distribution $\pi_{\theta}\parens{U}$. Each control trajectory is simulated (in parallel) to obtain a cost $J^{(i)}$, and all costs are jointly used to update the sampling parameters $\theta$. The control input $u(t)$ can be obtained for any time $t$ from the spline parameters $U$. This allows planning to proceed asynchronously, with the parameters $\theta$ updated as quickly as computational limits allow.

\SetKwComment{Comment}{// }{}
\begin{algorithm}[h]

\caption{Sampling-based Predictive Control}
\label{alg:sampling_planner}
\KwIn{$\theta$, $N$, planner-specific parameters.}

\While{planning}{
    $x_0 \leftarrow \hat{x}(t)$ \Comment*[r]{estimate curr state}
    \For{$i=1$ to $N$ \Comment*[r]{multi-threaded}}{
        $U^{(i)} \sim \pi_{\theta}(U)$ \Comment*[r]{sample controls}
        $J^{(i)} \gets J\left(U^{(i)}; x_0\right)$ \Comment*[r]{eval rollout}
    }
    $\theta \leftarrow \mathtt{update\_params}\parens{U^{(1:N)}, J^{(1:N)}}$;
    
    $u(t) \gets \mathtt{get\_action}(\theta, t)$ \Comment*[r]{asynchronous}
}
\end{algorithm}

In our experiments, we test two simple planning strategies that both use a diagonal Gaussian distribution $\pi_{\theta}\parens{U} = \mathcal{N}\parens{\widebar{U}, \Sigma}$ with parameters $\theta=\parens{\widebar{U}, \Sigma}$. In Algorithm~\ref{alg:sampling_planner}, $\mathtt{get\_action}(\theta, t)$ is a spline interpolation with knots $\widebar{U}$.

\textbf{Predictive sampling (PS)} repeatedly updates $\widebar{U}$ to be the best sample, fixing $\Sigma=\sigma^2 I$. Despite its simplicity, PS has demonstrated surprisingly effective performance on complex robot manipulation and locomotion tasks in simulation \cite{howell2022_mjpc}. 

The \textbf{cross-entropy method (CEM)} instead fits both $\widebar{U}$ and $\Sigma=\diag(s)$ to the sample mean and variance of the $M$ best \textit{elite samples}, where $s$ is a vector of covariances. CEM is only marginally more complex than PS, but has long been used for general gradient-free optimization\cite{rubinstein1999_cem}.

As in prior work \cite{handa2024_dextreme}, the task considered in this paper is to use a dexterous hand to rotate a cube to within 0.4 \unit{\radian} of as many goal orientations in a row as possible without dropping. The goals are uniformly randomly sampled over $SO(3)$. In contrast with \cite{handa2024_dextreme}, to ensure sufficient task difficulty, each new goal must be at least $90^\circ$ from the prior one.

Mathematically, the DROP cube reorientation problem is expressed as the optimal control problem
\eqs{\label{eqn:ocp}
    \min_u &\int_{0}^{T} \biggl\{ \lambda_\text{g} \cdot \ell_\text{g}(r^c(t)) + \lambda_\text{p} \cdot \ell_\text{p}(p^c(t)) \biggr\} dt \\
    \text{s.t.}&\quad \dot{x}(t) = f(x(t),u(t)), \\
    &\quad x(0) = x_0,
}
where $\lambda_{(\cdot)}$ denote weights, the dynamics $f(x, u)$ are only available through simulation, and
\eqa{
    \ell_\text{g}(r^c) :=& \norm{r^c \ominus r^c_\text{goal}}_2^2, \\
    \ell_\text{p}(p^c) :=& \text{dist}_\mathcal{S}(p^c), \label{eqn:pos_cost}
}
are running costs.
The variables $p^c$ and $r^c$ denote the positional and rotational components of the cube pose $q^c=\brackets{p^c, r^c}$, which are extracted from the simulated state $x(t)$. $\ell_\text{g}$ penalizes rotational distance from the goal, while $\ell_\text{p}$ penalizes the cube leaving a ``safe'' region $\mathcal{S}$ in Cartesian space.

In this work, we prioritize drop reduction by setting $\lambda_\text{p}$ high and choosing a conservative region $\mathcal{S}$ (for details, see \cite{li2024_drop_website}). In practice, we use a relatively low $\lambda_\text{g}$, which slows down the planner but also amplifies differences in rotation rate across methods, allowing easier quantitative comparison. 

\subsection{The Pose Estimation Pipeline}\label{sec:estimation}
Our pose estimator consists of three parts: a keypoint predictor, a fixed-lag smoother, and a collision-aware corrector.

\textbf{Keypoint Prediction.} The estimator takes in images $I_c\in\mathbb{R}^{C \times H \times W}$ from $n_c$ cameras. We first train a vision model $g_\varphi$ that predicts 8 fixed keypoints on the cube corresponding to its corners, $p_\text{kp}=g_\varphi(I_c)$. The keypoint prediction task is supervised from a training dataset of 686,000 images of a simulated cube rendered by \verb|Blender|, which includes ground-truth pixel locations for all keypoints, even those that are outside the frame or occluded. We generate randomized background scenes using \verb|kubric| \cite{greff2021kubric}, which spawns the cube along with other random assets in a \verb|pybullet| simulation. Similar datasets are commonly used to train ``track-any-point'' models \cite{doersch2024_bootstap, karaev2023_cotracker, xiao2024_spatialtracker} which exhibit strong sim-to-real transfer for similar keypoint tracking tasks.

\begin{figure}[t]
    \vspace{0.25cm}
    \centering
    \includegraphics[width=\linewidth]{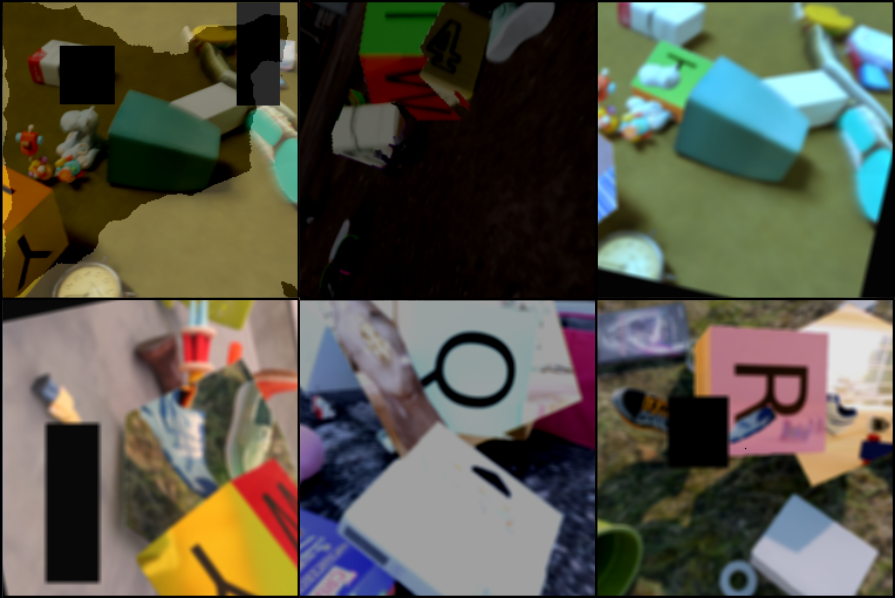}
    \caption{
        \textbf{Image augmentations.} To train the keypoint predictor, we augmented simulated images of the cube with random crops, affine transformations, spliced backgrounds, random deletions, and visual adjustments in color, contrast, brightness, and reflectivity.
    }
    \label{fig:augmentations}
    \vspace{-0.5cm}
\end{figure}

Crucially, we then augment these images with random affine transforms; visual effects like color, shadow, and contrast; and randomized backgrounds. We also found that pruning images where the cube was nearly occluded, or too close for a reliable pose estimate, was essential for good performance. Figure~\ref{fig:augmentations} shows some of the resulting images.

To train $g_\varphi$, we fine-tuned an ImageNet-pretrained \verb|resnet18| using \verb|AdamW| with an MSE loss. The only model adjustments were the number of input channels (depending on RGB or RGBD inputs) and the dimension of the output layer. For finer details, see \cite[Extended Version]{li2024_drop_website}.

\textbf{Pose Smoothing.} We use GTSAM \cite{dellaert2022_gtsam}, a factor graph-based state estimation package, to convert keypoints into a cube pose estimate via fixed-lag smoothing. Given a fixed pinhole camera model with known camera poses and cube size, we derive keypoint measurement factors that relate keypoints $p_\text{kp}$ to a cube pose $q^c$. This allows GTSAM to fuse keypoint predictions from an arbitrary number of cameras in real-time to yield a smoothed cube pose estimate $\tilde{q}^c$. See our open-source implementation for exact details \cite{li2024_drop_website}.

In both simulation and hardware, we estimate velocities by numerically differentiating position estimates (drawn from the smoother for the cube and joint encoders for the hand) and applying an exponential moving average filter with parameter $\alpha=0.1$ to compute a smooth velocity estimate $\tilde{v}$.

\textbf{Collision-Aware Correction.} While the smoother's cube pose estimate $\tilde{q}^c$ is usually accurate within 1\unit{\cm}, it may not always be physically compatible with the hand configuration $\tilde{q}^h$ from the encoders, as the smoother has no knowledge of collision dynamics. Thus, $\tilde{q}=[\tilde{q}^c, \tilde{q}^h]$ often corresponds to non-negligible interpenetration between cube and hand, which (i) leads to inaccurate plans that destabilize the closed-loop system, and (ii) decreases the planning rate, as stiff MuJoCo models with high interpenetration are poorly-conditioned, requiring more solver iterations.

To remedy this, we adapt the method of \cite{abouchakra2024_gswrenchcorrector} by using a \textit{corrector}, which maintains an asynchronous internal model 
\begin{equation}\label{eq:corrector_model}
    \dot{\hat{x}} = \hat{f}(\hat{x}(t), u(t))
\end{equation}
with state $\hat{x} = [\hat{q}^c, \hat{q}^h, \hat{v}^c, \hat{v}^h]$.

\begin{figure*}[t]
    \vspace{0.25cm}
    \centering
    \includegraphics[width=\linewidth]{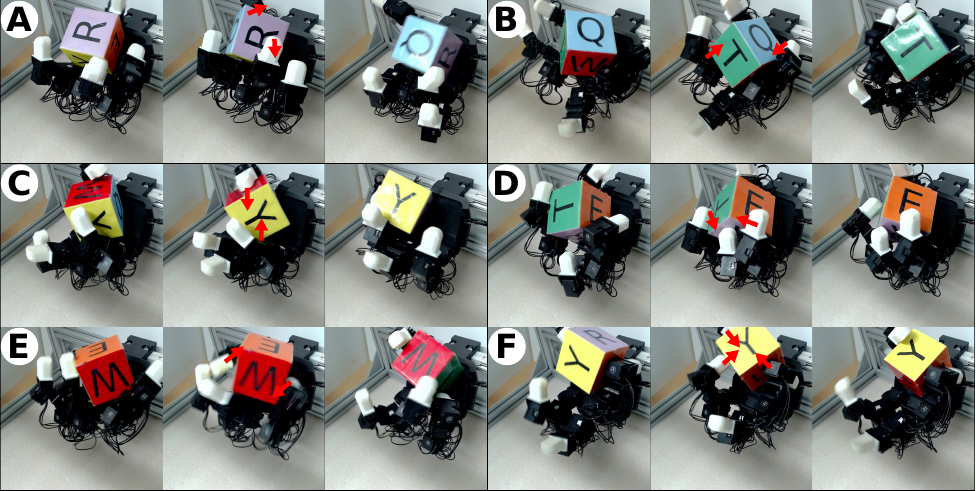}
    \caption{\textbf{Examples of rotations.} CEM can discover many contact-rich plans for cube reorientation. The red arrows show where forces are primarily applied to achieve rotations. \textbf{(A)} The middle finger pushes down on a cube edge while the base of the thumb lifts the opposite corner, rotating the Q face up. \textbf{(B)} The ring finger and base of the index finger push on opposite corners to rotate the T face up. \textbf{(C)} The thumb pulls down on the W face while the base of the ring finger pushes on the opposite corner to rotate the Y face up. \textbf{(D)} The index finger pushes down on the edge of the T face while the ring finger swipes left on the E face to rotate it up. \textbf{(E)} The ring finger first swipes inwards, then the thumb quickly follows to pull the W face up. \textbf{(F)} The thumb and the ring finger pinch and lift the cube, then the index finger pushes on an edge to rotate the Y face up. The cube is calmly lowered onto the palm.}
    \vspace{-0.5cm}
    \label{fig:rotation_examples}
\end{figure*}

The corrector receives an un-adjusted estimate $\tilde{x}$ from the smoother and encoders, and computes a ``corrective wrench''
\begin{equation}\label{eqn:corrector_wrench}
    w = - C_P(\hat{q}^c \ominus \tilde{q}^c) - C_D(\hat{v}^c - \tilde{v}^c)
\end{equation}
that is added to \eqref{eq:corrector_model} as a generalized force. Since the corrector starts in a feasible state, simulating \eqref{eq:corrector_model} results in a corrected estimate $\hat{x}$ that is always physically feasible, but pulled toward the (possibly infeasible) vision-based estimate $\tilde{x}$. We also add a constant corrective force in the direction of gravity to counteract any smoother estimates that may unrealistically pull the cube upwards due to high corrector gains $C_P$, overwhelming the natural gravitational forces of \eqref{eq:corrector_model}.

\section{Experiments}\label{sec:experiments}

For the remainder of the paper, we thoroughly evaluate the DROP architecture via hardware and simulation experiments designed to answer the following questions:
\begin{enumerate}[noitemsep,topsep=0pt,parsep=0pt,partopsep=0pt]
    \item Can DROP perform robust and dynamic cube reorientation in hardware (Sec. \ref{sec:hardware})?
    \item How do different components of DROP  contribute to its overall performance and reliability (Sec. \ref{sec:hardware_ablations})?
    \item How do modeling and state estimation error affect DROP's performance (Sec. \ref{sec:simulation})?
\end{enumerate}

Overall, we find that DROP can reliably perform cube reorientation in hardware, achieving performance comparable with RL-based methods while exhibiting surprising robustness and dexterity. We emphasize that the goal of our experiments is not to design the best-possible control strategy, but to assess the viability of sampling-based online planning for the cube reorientation task via thorough evaluations of the DROP architecture. Thus, we leave systematic comparisons of DROP with other methods for future work.

\subsection{Hardware Setup and Experimental Details}\label{sec:experimental_setup}
For all experiments, we perform in-hand rotation of a 3D-printed cube with 7\unit{\cm} side lengths and a fixed-base LEAP hand \cite{shaw2023_leap} with palm angled downwards at 20$^\circ$ so that the cube slides off without intervention (see Fig. \ref{fig:hero} and \ref{fig:rotation_examples}). We use a 128-thread Ryzen Threadripper Pro 5995wx CPU to plan rollouts in parallel. To capture images, we use three ZED Mini cameras with RGBD channels and perform keypoint estimation on 256x256 center-cropped images at VGA resolution. The estimator runs at about 90\unit{\Hz}, while the planner frequency fluctuates between 25-50\unit{\Hz}.

For our trials, we compare three planners: PS and CEM (the two sampling-based methods discussed in Sec. \ref{sec:control}), as well as iLQR \cite{li2004_ilqr}, a gradient-based method. PS and CEM use $N=120$ rollouts while iLQR devotes 120 threads to parallel line search. CEM uses $M=4$ elite samples. When tuning hyperparameters (such as cost weights), we prioritize robustness at the expense of rotation speed. For all experiments, we used identical costs, code, and hyperparameters for performing simulation rollouts, state estimation, and communicating with hardware. Unlike previous work \cite{openai2018_dactyl, handa2024_dextreme}, we did not observe any significant degradation of our hardware stack (e.g., overheating) during testing.

\begin{table*}[t]
\vspace{0.25cm}

\centering
\setlength{\tabcolsep}{4pt}
\renewcommand{\arraystretch}{1.1}

\begin{tabular}{|c|c||c|c|c||c|c|}
\hline
& \textbf{Planner} & \textbf{Num Rots (Sorted)} & \textbf{Mean Rots} & \textbf{Median Rots} & \textbf{Mean Rot/\unit{\s}} $\uparrow$ & \textbf{Median Rot/\unit{\s}} $\uparrow$ \\ \hline \hline

\multirow{2}{*}{\parbox[t]{2.25cm}{\raggedright \textbf{(A)} Prior Work}} & Dactyl \cite{openai2018_dactyl} & 9, 12, 13, 19, 28, 29, 29, 32, 43, 50 & $26.4 \pm 13.4$ & 28.5 & Not Reported & Not Reported \\ \cline{2-7}
& Dextreme \cite{handa2024_dextreme} & 1, 6, 6, 10, 10, 18, 18, 36, 61, 112 & $27.8 \pm 19.0$ & 14.0 & Not Reported & Not Reported \\ \hline \hline

\multirow{4}{*}{\parbox[t]{2.25cm}{\raggedright \textbf{(B)} Main Results}} & iLQR & 0, 0, 0, 0, 0, 0, 0, 0, 0, 0 & $0 \pm 0$ & 0 & N/A & N/A \\ \cline{2-7}
& PS & 0, 0, 1, 1, 1, 2, 2, 3, 4, 5 & $1.9 \pm 1.58$ & 1.5 & $0.063 \pm 0.070$ & 0.033 (0.021, 0.061) \\ \cline{2-7}
& CEM & 2, 2, 11, 11, 17, 22, 27, 34, 39, 48 & $21.3 \pm 14.8$ & 19.5 & $0.090 \pm 0.081$ & 0.062 (0.038, 0.106) \\ \cline{2-7}
& CEM (No T/O) & 2, 10, 11, 11, 22, 39, 48, 53, 59, 81 & $33.6 \pm 24.9$ & 30.5 & $0.086 \pm 0.085$ & 0.063 (0.038, 0.103) \\ \hline \hline

\multirow{3}{*}{\parbox[t]{2.25cm}{\raggedright \shortstack[c]{\textbf{(C)} CEM Ablations \\ (No T/O)}}} & RGB Only & 1, 1, 12, 13, 20, 37, 52, 79, 94, 128 & $43.7 \pm 41.4$ & 28.5 & $0.081 \pm 0.106$ & 0.053 (0.030, 0.089) \\ \cline{2-7}
& No Corrector & 3, 4, 6, 9, 11, 14, 17, 20, 40, 97 & $22.1 \pm 27.0$ & 12.5 & $0.065 \pm 0.063$ & 0.046 (0.029, 0.080) \\ \cline{2-7}
& Half Rollouts & 2, 7, 17, 21, 21, 27, 29, 37, 44, 65 & $27.0 \pm 17.4$ & 24.0 & $0.062 \pm 0.070$ & 0.040 (0.022, 0.072) \\ \hline
\end{tabular}

\caption{
    \textbf{Hardware experiments.} \textbf{(A)} Prior works using RL for cube reorientation. Note that \cite{openai2018_dactyl} performs \textit{axis-aligned} rotations. Best results are shown. \textbf{(B)} Among online planners, CEM clearly performs the best. All planners use 120 threads, RGBD images, and the cube pose corrector. We note that when ignoring the 80\unit{\s} timeout imposed in \cite{openai2018_dactyl, handa2024_dextreme}, CEM achieves higher mean and median rotation counts than Dactyl and Dextreme. \textbf{(C)} Hardware ablations show that using depth images slightly improves rotation rate, but using the corrector and as many threads as possible substantially boosts performance.
}
\label{tab:summary}
\vspace{-0.5cm}
\end{table*}

\subsection{Main Hardware Results}\label{sec:hardware}

We begin by presenting results of hardware trials using the full stack shown in Fig. \ref{fig:hero} and discussed in Sec. \ref{sec:method}. We study each planner by running 10 trials of the cube reorientation task and analyzing the associated rotation and timing statistics. For examples of interesting rotations, see Fig. \ref{fig:rotation_examples}, and for quantitative results, see Table \ref{tab:summary}B. Following prior work, we report statistics assuming the run ends if 80\unit{\s} have elapsed without reaching a goal. Since this cutoff is arbitrary and we used a fairly conservative cost that slows rotation rate, we also report results for the same runs while ignoring the timeout period, ending only when the cube is dropped. For iLQR and PS, this did not change the results.

\textbf{CEM greatly outperforms PS and iLQR.} While iLQR is unable to achieve any rotations and PS only achieves single-digit rotations in hardware, CEM is able to achieve dozens of rotations. Moreover, the rate that it can rotate the cube is also nearly 50\% faster than PS, suggesting that it can discover and/or execute contact-rich plans much more effectively.

\textbf{CEM discovers nontrivial contact sequences.} CEM finds contact-rich plans that leverage contacts with all parts of the hand. Many rotations are only feasible when the hand first partially rotates the cube using an initial contact sequence, then completes the rotation by gaiting the cube to a different set of contacts. For example, in Fig. \ref{fig:rotation_examples}D, the cube is first pushed forward and continually supported by the thumb, allowing the index and ring fingers to then swipe in opposite directions to rotate the cube. Similarly, in Fig. \ref{fig:rotation_examples}F, the thumb and ring finger first pinch and lift the entire cube, which allows the index finger to rotate it about the pinched axis before the cube is safely lowered back onto the palm.

Moreover, many discovered plans exploit contact modes that are traditionally challenging to find, like sliding on edges and corners. To execute these motions, the planner employs intuitive strategies like maximizing torque by levering the cube close to a corner or edge. Lastly, our conservative cost function also induced safeguarding behavior, where fingers preemptively blocked the cube from the palm's edges or carefully supported the cube during risky rotations.

\textbf{The gradient-based iLQR planner is not viable.} Corroborating recent work \cite{howell2022_mjpc, suh2022_diffsim}, we find that the stiff dynamics  of the cube reorientation task prevent gradient-based methods from effectively finding good plans most likely due to poor numerical conditioning, causing jerky, erratic behaviors that do not lead to coherent rotation sequences.

\textbf{DROP performs comparably to RL.} While it is challenging to compare our results to prior RL-based methods like Dactyl \cite{openai2018_dactyl} or Dextreme \cite{handa2024_dextreme} due to many factors distinguishing each setup, like hand morphology, cube size, physical properties, camera type, or vision model, Table \ref{tab:summary}A/B shows that our simple CEM planner approaches the performance of RL-based rotation policies (outperforming them when ignoring timeouts) with similar dexterity (see Fig. \ref{fig:rotation_examples}).

\begin{figure}
    \centering
    \includegraphics[width=\linewidth]{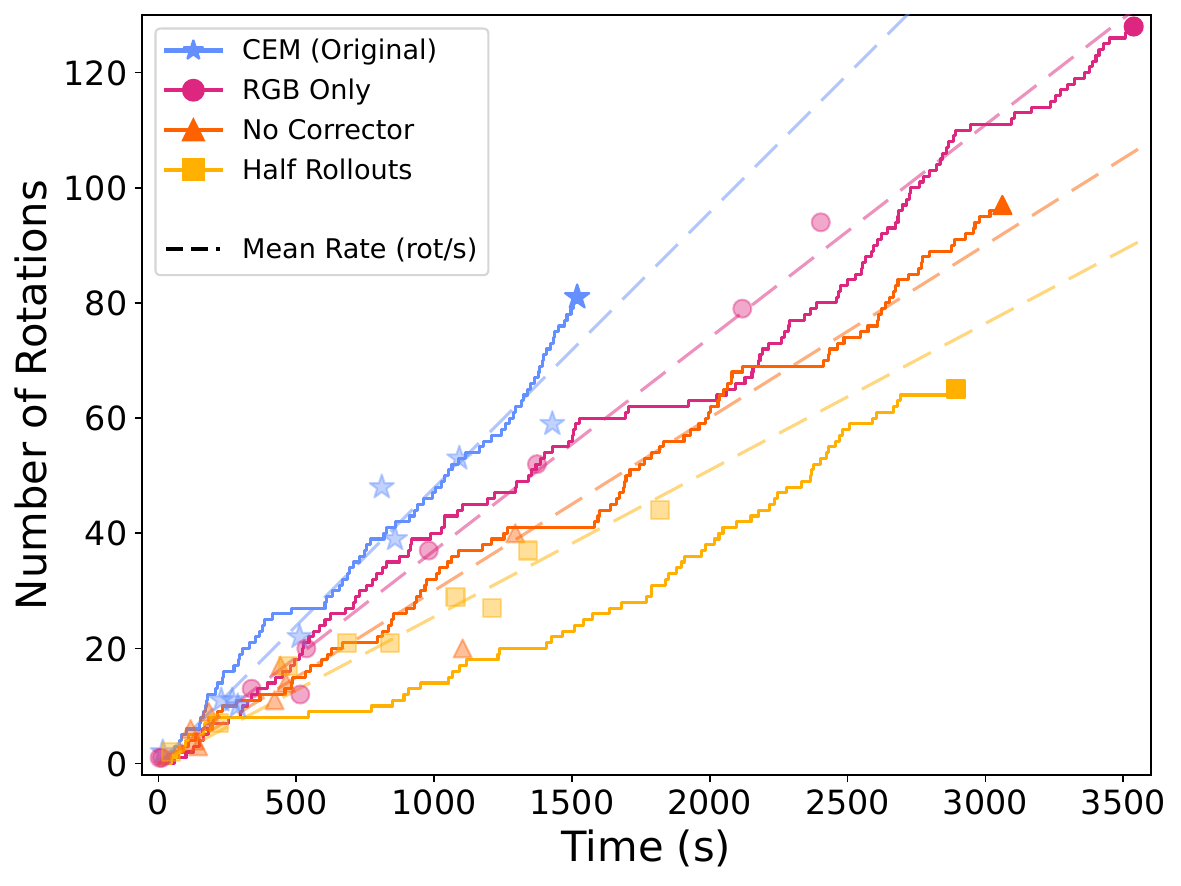}
    \caption{
        \textbf{CEM ablation rotation rates.} We use rotation rate as a proxy for planner robustness, as slower rates correspond to ``stuck'' plans or repeatedly failed moves. Markers are all rotations vs. times for CEM and all ablations. The dashed lines show the mean rotation rates: it is clear that all ablations decrease the rate, which justifies our design of the DROP architecture. The solid lines show individual rotations for each method's longest streak. The long, flat regions correspond to the planner getting ``stuck'' in local minima.
    }
    \label{fig:ablation_speeds}
    \vspace{-0.75cm}
\end{figure}

\subsection{Hardware Ablations}\label{sec:hardware_ablations}

To understand the impact of key design choices in the DROP architecture, we conducted a series of single-variable ablations with the CEM planner on hardware (Table \ref{tab:summary}C).

\textbf{Depth improves performance, but only marginally.} We compared keypoint detection models trained on RGB versus RGBD images. While RGBD slightly improved rotation rates, it did not significantly outperform RGB, which still achieved 128 rotations, the longest sequence ever observed for DROP. Despite this noisy result, RGB's median rotations (28.5) were still slightly lower than RGBD's (30.5), and overall, the relative rotation rates suggest that depth provides a minor improvement in state estimation accuracy.

\textbf{The corrector is key for performance.} We tested DROP when ablating the corrector, passing raw smoother estimates $\tilde{q}$ directly to the planner. This significantly reduced performance, with mean rotations decreasing by 33\%, median rotations by 60\%, and rotation rate by 25\%. Without the corrector, the planner often became trapped in local minima and long periods of inactivity (flat regions of rollouts in Fig. \ref{fig:ablation_speeds}). This was caused by the exploitation of non-physical forces in rollouts arising from non-physical hand-cube interpenetration, leading to unrealistic predictions.

\textbf{DROP is sensitive to the number of rollouts.} We reduced the number of rollouts from 120 to 60 and elite samples from 4 to 3, resulting in a 20\% decrease in mean rotation count and an over 25\% reduction in rotation rate. This highlights the importance of variance reduction via sufficiently-high sample quantity in SPC, and suggesting that improving search efficiency could significantly boost performance.

\textbf{All ablations increased rotation rate variance.} This consistent pattern demonstrates that depth measurements, the corrector, and sufficient rollout quantity all contribute significantly to the planner's reliability and consistency, which justifies the design of the DROP architecture.

\subsection{Simulated Robustness Study}\label{sec:simulation}

Lastly, we study DROP's robustness to model and estimation errors by conducting controlled trials in simulation, letting us isolate the planner from the estimation pipeline. We compared iLQR, PS, and CEM under various corrupted conditions, simulating system physics with a 2\unit{\ms} timestep while planner threads utilized a separate physics model with a 10\unit{\ms} timestep. Each configuration underwent five trials, ending upon cube drop, 80\unit{\s} timeout, or 150 rotations.

Specifically, we intentionally induce two types of errors that we believe contribute to real-world brittleness: (i) mis-tuning the planners' internal value of the hand $K_p$ gains, and (ii) corrupting pose estimates from the simulation with a 0.1\unit{\s} lag and additive noise (simulated using a bounded random walk to mimic asymmetric state estimation error). When corrupting $K_p$, we study two cases: multiplying the true value by 1.25x and 1.5x, as the results were enlightening for comparing different planners. We also study the effect of the estimator and $K_p$ errors together.

\textbf{CEM is the most robust planner.} Table \ref{tab:sim} shows that CEM is the best planner under all error conditions. For example, while PS achieves a higher mean rotation count under perfect conditions than CEM, when $K_p$ is mildly corrupted up to 1.25x, PS immediately achieves fewer mean rotations than CEM while suffering an over 2x decrease in rotation rate. At 1.5x, PS hardly rotates the cube at all, while CEM achieves 32.2 mean rotations, and even with the most aggressive errors, CEM was able to achieve dozens of rotations. CEM's superiority can be attributed to its strategy of recomputing $\pi_\theta$ from multiple rollouts, in contrast to PS's single-rollout approach. This strategy appears to be key for robustness, providing a plausible explanation for CEM's markedly better performance in hardware.

\textbf{Rot/\unit{\s} is a good proxy for robustness.} Based on empirical observation, we find that low rotation rates are typically caused by the failure to execute precisely-planned motions or the inability of the planner to escape local minima, resulting in the cube being ``stuck,'' which can be attributed to model or estimation error. CEM's superior speed in these simulations supports our assessments of its robustness on hardware.

\textbf{Error types have distinct failure modes.} While incorrect $K_p$ values primarily resulted in timeouts, corrupted estimates typically caused cube drops. This suggests that perfect state estimates allow for safe "caging" even with poor actuation models, but estimation errors during precise maneuvers often cause drops, as the planner underestimates rollout risk.
\begin{table}[t]
\vspace{0.25cm}

\centering
\setlength{\tabcolsep}{1pt}
\renewcommand{\arraystretch}{1.2}  

\begin{tabular}{|c|c||c|c||c|c|}
\hline
\textbf{Planner} & \textbf{Change} & \textbf{Mean Rot/\unit{\second} $\uparrow$} & \textbf{Mean Rots} & \textbf{Drops} & \textbf{Timeouts} \\
\hline
\hline

iLQR & None & N/A & $0 \pm 0$ & 3 & 2\\  

\hline

\multirow{5}{*}{PS} & None & 0.130 & $122 \pm 54.8$ & 1 & 0 \\  
& $K_p$ x1.25 & 0.060 & $44.4 \pm 26.4$ & 1 & 4 \\  
& $K_p$ x1.5 & 0.042 & $1.2 \pm 1.17$ & 0 & 5 \\  
& Est. & 0.032 &$3.2 \pm 1.72$ & 5 & 0 \\  
& Est., $K_p$ x1.25 & 0.048 & $3.4 \pm 3.14$ & 5 & 0 \\  
& Est., $K_p$ x1.5 & 0.015 & $0.2 \pm 0.40$ & 3 & 2 \\

\hline

\multirow{5}{*}{CEM} & None & 0.222 & $95.4 \pm 34.8$ & 0 & 4 \\  
& $K_p$ x1.25 & 0.148 & $52.8 \pm 26.0$ & 0 & 5 \\  
& $K_p$ x1.5 & 0.100 & $32.2 \pm 26.6$ & 0 & 5 \\  
& Est. & 0.120 & $46.8 \pm 24.7$ & 5 & 0 \\  
& Est., $K_p$ x1.25 & 0.083 & $73.8 \pm 26.4$ & 4 & 1 \\  
& Est., $K_p$ x1.5 & 0.058 & $24.8 \pm 13.3$ & 2 & 3 \\
\hline

\end{tabular}

\caption{
    \textbf{Simulated robustness tests.} We intentionally degrade planners to test their robustness by (i) tuning the hand proportional gain $K_p$ too high, and (ii) corrupting the estimator with noise and lag (denoted ``Est.''). iLQR failed to achieve any rotations even with perfect information. We observe that PS degrades substantially more than CEM in the presence of model and estimation error, which suggests that CEM may transfer well to hardware.
}
\vspace{-0.5cm}
\label{tab:sim}
\end{table}

\section{Conclusion and Future Directions}\label{sec:conclusion}
This work presents DROP, a minimalist online planning method for in-hand manipulation via sampling-based predictive control that achieves robust cube rotations in hardware. While promising, there are many avenues for future research.

\textbf{Better planners.} While we found that vanilla CEM already achieved impressive results, many more sophisticated algorithms exist, such as CMA-ES \cite{jankowski2023_vpsto}, MPPI \cite{williams2018_mppi}, etc.

\textbf{Robustness.} DROP often plans ``risky'' actions, possibly due to differences between real-world and simulated physics. Incorporating domain randomization or risk-sensitivity into search-based planners, like successful RL approaches, remains an open challenge, especially due to the extra computation required to simulate randomized physics online. 

\textbf{Object generality.} As in prior RL-based works \cite{openai2018_dactyl, handa2024_dextreme}, we first focus on robustly reorienting a single, simple object. While DROP can easily adapt to new objects in simulation, our current vision pipeline requires retraining for new objects. Recent advancements in general pixel-space tracking \cite{doersch2024_bootstap, xiao2024_spatialtracker} and video-based mask propagation \cite{ravi2024_sam2} suggest more avenues for pose estimation that could generalize our search-based approach without extensive retraining.

\textbf{Enhanced, data-driven search.} As noted in our ablations, finding good plans via search demands many threads; indeed, our work relies on a server-grade CPU, since dynamics simulation is about an order of magnitude slower on GPUs. Thus, improving efficiency is key for better performance. Promising directions include sampling from imitation-learned policies \cite{chi2024_diffusion}, learning value functions for rollout evaluation \cite{silver2016_mastering}, and exploring alternate spline parameterizations \cite{jankowski2023_vpsto} or action spaces \cite{martin2019_iros}. Searching for high-level commands for a lower-level RL policy could perhaps yield systems with both the flexibility of search and robustness of RL.

DROP opens many paths for contact-rich manipulation. It is our hope that algorithms like DROP can generalize to more real-world tasks than cube reorientation, unlocking tool use, enhanced human-robot collaboration, and more.

\bibliographystyle{unsrt}
\bibliography{references}

\begin{thebibliography}{1}

\bibitem{sacks2023_dmpo}
Jacob Sacks, Rwik Rana, Kevin Huang, Alex Spitzer, Guanya Shi, and Byron Boots.
\newblock Deep model predictive optimization, 2023.

\bibitem{dellaert2017_factorgraphs}
Frank Dellaert and Michael Kaess.
\newblock {\em Factor Graphs for Robot Perception}, volume~6.
\newblock Foundations and Trends in Robotics, 2017.

\bibitem{barfoot2024state}
Timothy~D Barfoot.
\newblock {\em State estimation for robotics}.
\newblock Cambridge University Press, 2024.

\bibitem{paszke2019_pytorch}
Adam Paszke, Sam Gross, Francisco Massa, Adam Lerer, James Bradbury, Gregory Chanan, Trevor Killeen, Zeming Lin, Natalia Gimelshein, Luca Antiga, Alban Desmaison, Andreas Köpf, Edward Yang, Zach DeVito, Martin Raison, Alykhan Tejani, Sasank Chilamkurthy, Benoit Steiner, Lu~Fang, Junjie Bai, and Soumith Chintala.
\newblock Pytorch: An imperative style, high-performance deep learning library, 2019.

\bibitem{coumans2021_pybullet}
Erwin Coumans and Yunfei Bai.
\newblock Pybullet, a python module for physics simulation for games, robotics and machine learning.
\newblock \url{http://pybullet.org}, 2016--2021.

\end{thebibliography}


\begin{thebibliography}{10}

\bibitem{ishihara_dynamic2006}
Tatsuya Ishihara, Akio Namiki, Masatoshi Ishikawa, and Makoto Shimojo.
\newblock Dynamic pen spinning using a high-speed multifingered hand with high-speed tactile sensor.
\newblock In {\em 2006 6th IEEE-RAS International Conference on Humanoid Robots}, pages 258--263, 2006.

\bibitem{chavan_extrinsic2014}
Nikhil Chavan-Dafle, Alberto Rodriguez, Robert Paolini, Bowei Tang, Siddhartha Srinivasa, Michael Erdmann, Matthew~T. Mason, Ivan Lundberg, Harald Staab, and Thomas Fuhlbrigge.
\newblock Extrinsic dexterity: In-hand manipulation with external forces.
\newblock In {\em Proceedings of (ICRA) International Conference on Robotics and Automation}, pages 1578 -- 1585, May 2014.

\bibitem{openai2018_dactyl}
OpenAI, Marcin Andrychowicz, Bowen Baker, Maciek Chociej, Rafal J{\'{o}}zefowicz, Bob McGrew, Jakub Pachocki, Arthur Petron, Matthias Plappert, Glenn Powell, Alex Ray, Jonas Schneider, Szymon Sidor, Josh Tobin, Peter Welinder, Lilian Weng, and Wojciech Zaremba.
\newblock Learning dexterous in-hand manipulation.
\newblock {\em CoRR}, abs/1808.00177, 2018.

\bibitem{handa2024_dextreme}
Ankur Handa, Arthur Allshire, Viktor Makoviychuk, Aleksei Petrenko, Ritvik Singh, Jingzhou Liu, Denys Makoviichuk, Karl~Van Wyk, Alexander Zhurkevich, Balakumar Sundaralingam, Yashraj Narang, Jean-Francois Lafleche, Dieter Fox, and Gavriel State.
\newblock Dextreme: Transfer of agile in-hand manipulation from simulation to reality, 2024.

\bibitem{chen2023_reorientation}
Tao Chen, Megha Tippur, Siyang Wu, Vikash Kumar, Edward Adelson, and Pulkit Agrawal.
\newblock Visual dexterity: In-hand reorientation of novel and complex object shapes.
\newblock {\em Science Robotics}, 8(84), November 2023.

\bibitem{howell2022_mjpc}
Taylor Howell, Nimrod Gileadi, Saran Tunyasuvunakool, Kevin Zakka, Tom Erez, and Yuval Tassa.
\newblock Predictive sampling: Real-time behaviour synthesis with mujoco, 2022.

\bibitem{li2024_drop_website}
Albert Li.
\newblock Drop: Dexterous reorientation via online planning.
\newblock \url{https://caltech-amber.github.io/drop/}, 2024.

\bibitem{yin2023_rotating}
Zhao-Heng Yin, Binghao Huang, Yuzhe Qin, Qifeng Chen, and Xiaolong Wang.
\newblock Rotating without seeing: Towards in-hand dexterity through touch.
\newblock {\em Robotics: Science and Systems}, 2023.

\bibitem{rostel2023estimator}
Lennart R{\"o}stel, Johannes Pitz, Leon Sievers, and Berthold B{\"a}uml.
\newblock Estimator-coupled reinforcement learning for robust purely tactile in-hand manipulation.
\newblock In {\em 2023 IEEE-RAS 22nd International Conference on Humanoid Robots (Humanoids)}, pages 1--8. IEEE, 2023.

\bibitem{chen2021_generalreorientation}
Tao Chen, Jie Xu, and Pulkit Agrawal.
\newblock A system for general in-hand object re-orientation, 2021.

\bibitem{posa2014_direct_cito}
Michael Posa, Cecilia Cantu, and Russ Tedrake.
\newblock A direct method for trajectory optimization of rigid bodies through contact.
\newblock {\em The International Journal of Robotics Research}, 33:69 -- 81, 2014.

\bibitem{doshi_cito_locomotion}
Neel Doshi, Kaushik Jayaram, Benjamin Goldberg, Zachary Manchester, Robert Wood, and Scott Kuindersma.
\newblock Contact-implicit optimization of locomotion trajectories for a quadrupedal microrobot.
\newblock In {\em Robotics: Science and Systems XIV}, RSS2018. Robotics: Science and Systems Foundation, June 2018.

\bibitem{lecleach2023_cimpc}
Simon~Le Cleac'h, Taylor Howell, Shuo Yang, Chi-Yen Lee, John Zhang, Arun Bishop, Mac Schwager, and Zachary Manchester.
\newblock Fast contact-implicit model-predictive control, 2023.

\bibitem{kurtz2023_idto}
Vince Kurtz, Alejandro Castro, Aykut Özgün Önol, and Hai Lin.
\newblock Inverse dynamics trajectory optimization for contact-implicit model predictive control, 2023.

\bibitem{aydinoglu2022_c3}
Alp Aydinoglu and Michael Posa.
\newblock Real-time multi-contact model predictive control via admm, 2022.

\bibitem{yang2024_dynamic}
William Yang and Michael Posa.
\newblock Dynamic on-palm manipulation via controlled sliding.
\newblock In {\em Proceedings of Robotics: Science and Systems}, July 2024.

\bibitem{yang2024_contactsdf}
Wen Yang and Wanxin Jin.
\newblock Contactsdf: Signed distance functions as multi-contact models for dexterous manipulation, 2024.

\bibitem{chen2021_trajectotree}
Claire Chen, Preston Culbertson, Marion Lepert, Mac Schwager, and Jeannette Bohg.
\newblock Trajectotree: Trajectory optimization meets tree search for planning multi-contact dexterous manipulation.
\newblock In {\em 2021 IEEE/RSJ International Conference on Intelligent Robots and Systems (IROS)}, pages 8262--8268, 2021.

\bibitem{pang2023_quasistatic}
Tao Pang, H.~J.~Terry Suh, Lujie Yang, and Russ Tedrake.
\newblock Global planning for contact-rich manipulation via local smoothing of quasi-dynamic contact models, 2023.

\bibitem{lavalle2000_rrt}
Steven Lavalle and James Kuffner.
\newblock Rapidly-exploring random trees: Progress and prospects.
\newblock {\em Algorithmic and computational robotics: New directions}, 01 2000.

\bibitem{kavraki1996_prm}
L.E. Kavraki, P.~Svestka, J.-C. Latombe, and M.H. Overmars.
\newblock Probabilistic roadmaps for path planning in high-dimensional configuration spaces.
\newblock {\em IEEE Transactions on Robotics and Automation}, 12(4):566--580, 1996.

\bibitem{kalakrishnan2011_stomp}
Mrinal Kalakrishnan, Sachin Chitta, Evangelos Theodorou, Peter Pastor, and Stefan Schaal.
\newblock Stomp: Stochastic trajectory optimization for motion planning.
\newblock In {\em 2011 IEEE International Conference on Robotics and Automation}, pages 4569--4574, 2011.

\bibitem{williams2018_mppi}
Grady Williams, Paul Drews, Brian Goldfain, James~M. Rehg, and Evangelos~A. Theodorou.
\newblock Information-theoretic model predictive control: Theory and applications to autonomous driving.
\newblock {\em IEEE Transactions on Robotics}, 34(6):1603--1622, 2018.

\bibitem{jankowski2023_vpsto}
Julius Jankowski, Lara Brudermüller, Nick Hawes, and Sylvain Calinon.
\newblock Vp-sto: Via-point-based stochastic trajectory optimization for reactive robot behavior.
\newblock In {\em 2023 IEEE International Conference on Robotics and Automation (ICRA)}, pages 10125--10131, 2023.

\bibitem{pezzato2023_isaac_mppi}
Corrado Pezzato, Chadi Salmi, Max Spahn, Elia Trevisan, Javier Alonso-Mora, and Carlos~Hernandez Corbato.
\newblock Sampling-based model predictive control leveraging parallelizable physics simulations, 2023.

\bibitem{bhardwaj2021_storm}
Mohak Bhardwaj, Balakumar Sundaralingam, Arsalan Mousavian, Nathan~D. Ratliff, Dieter Fox, Fabio Ramos, and Byron Boots.
\newblock {STORM}: An integrated framework for fast joint-space model-predictive control for reactive manipulation.
\newblock In {\em 5th Annual Conference on Robot Learning}, 2021.

\bibitem{alvarezpadilla2024realtimewholebodycontrollegged}
Juan Alvarez-Padilla, John~Z. Zhang, Sofia Kwok, John~M. Dolan, and Zachary Manchester.
\newblock Real-time whole-body control of legged robots with model-predictive path integral control, 2024.

\bibitem{doersch2024_bootstap}
Carl Doersch, Pauline Luc, Yi~Yang, Dilara Gokay, Skanda Koppula, Ankush Gupta, Joseph Heyward, Ignacio Rocco, Ross Goroshin, João Carreira, and Andrew Zisserman.
\newblock Bootstap: Bootstrapped training for tracking-any-point, 2024.

\bibitem{karaev2023_cotracker}
Nikita Karaev, Ignacio Rocco, Benjamin Graham, Natalia Neverova, Andrea Vedaldi, and Christian Rupprecht.
\newblock {CoTracker}: It is better to track together.
\newblock 2023.

\bibitem{xiao2024_spatialtracker}
Yuxi Xiao, Qianqian Wang, Shangzhan Zhang, Nan Xue, Sida Peng, Yujun Shen, and Xiaowei Zhou.
\newblock Spatialtracker: Tracking any 2d pixels in 3d space.
\newblock In {\em Proceedings of the IEEE/CVF Conference on Computer Vision and Pattern Recognition (CVPR)}, 2024.

\bibitem{wen2024_foundationpose}
Bowen Wen, Wei Yang, Jan Kautz, and Stan Birchfield.
\newblock {FoundationPose}: Unified 6d pose estimation and tracking of novel objects.
\newblock In {\em CVPR}, 2024.

\bibitem{todorov2012mujoco}
Emanuel Todorov, Tom Erez, and Yuval Tassa.
\newblock Mujoco: A physics engine for model-based control.
\newblock In {\em 2012 IEEE/RSJ international conference on intelligent robots and systems}, pages 5026--5033. IEEE, 2012.

\bibitem{rubinstein1999_cem}
Reuven Rubinstein.
\newblock The cross-entropy method for combinatorial and continuous optimization, 1999.

\bibitem{greff2021kubric}
Klaus Greff, Francois Belletti, Lucas Beyer, Carl Doersch, Yilun Du, Daniel Duckworth, David~J Fleet, Dan Gnanapragasam, Florian Golemo, Charles Herrmann, Thomas Kipf, Abhijit Kundu, Dmitry Lagun, Issam Laradji, Hsueh-Ti~(Derek) Liu, Henning Meyer, Yishu Miao, Derek Nowrouzezahrai, Cengiz Oztireli, Etienne Pot, Noha Radwan, Daniel Rebain, Sara Sabour, Mehdi S.~M. Sajjadi, Matan Sela, Vincent Sitzmann, Austin Stone, Deqing Sun, Suhani Vora, Ziyu Wang, Tianhao Wu, Kwang~Moo Yi, Fangcheng Zhong, and Andrea Tagliasacchi.
\newblock Kubric: a scalable dataset generator.
\newblock 2022.

\bibitem{dellaert2022_gtsam}
Frank Dellaert and GTSAM Contributors.
\newblock borglab/gtsam, May 2022.

\bibitem{abouchakra2024_gswrenchcorrector}
Jad Abou-Chakra, Krishan Rana, Feras Dayoub, and Niko Sünderhauf.
\newblock Physically embodied gaussian splatting: A realtime correctable world model for robotics, 2024.

\bibitem{shaw2023_leap}
Kenneth Shaw, Ananye Agarwal, and Deepak Pathak.
\newblock Leap hand: Low-cost, efficient, and anthropomorphic hand for robot learning, 2023.

\bibitem{li2004_ilqr}
Weiwei Li and Emanuel Todorov.
\newblock Iterative linear quadratic regulator design for nonlinear biological movement systems.
\newblock In {\em Proceedings of the First International Conference on Informatics in Control, Automation and Robotics - Volume 1: ICINCO,}, pages 222--229. INSTICC, SciTePress, 2004.

\bibitem{suh2022_diffsim}
H.~J.~Terry Suh, Max Simchowitz, Kaiqing Zhang, and Russ Tedrake.
\newblock Do differentiable simulators give better policy gradients?, 2022.

\bibitem{ravi2024_sam2}
Nikhila Ravi, Valentin Gabeur, Yuan-Ting Hu, Ronghang Hu, Chaitanya Ryali, Tengyu Ma, Haitham Khedr, Roman R{\"a}dle, Chloe Rolland, Laura Gustafson, Eric Mintun, Junting Pan, Kalyan~Vasudev Alwala, Nicolas Carion, Chao-Yuan Wu, Ross Girshick, Piotr Doll{\'a}r, and Christoph Feichtenhofer.
\newblock Sam 2: Segment anything in images and videos.
\newblock {\em arXiv preprint arXiv:2408.00714}, 2024.

\bibitem{chi2024_diffusion}
Cheng Chi, Zhenjia Xu, Siyuan Feng, Eric Cousineau, Yilun Du, Benjamin Burchfiel, Russ Tedrake, and Shuran Song.
\newblock Diffusion policy: Visuomotor policy learning via action diffusion.
\newblock {\em The International Journal of Robotics Research}, 2024.

\bibitem{silver2016_mastering}
David Silver, Aja Huang, Chris~J. Maddison, Arthur Guez, Laurent Sifre, George van~den Driessche, Julian Schrittwieser, Ioannis Antonoglou, Veda Panneershelvam, Marc Lanctot, Sander Dieleman, Dominik Grewe, John Nham, Nal Kalchbrenner, Ilya Sutskever, Timothy Lillicrap, Madeleine Leach, Koray Kavukcuoglu, Thore Graepel, and Demis Hassabis.
\newblock Mastering the game of go with deep neural networks and tree search.
\newblock {\em Nature}, 529(7587):484–489, January 2016.

\bibitem{martin2019_iros}
Roberto Mart\'in-Mart\'in, Michelle Lee, Rachel Gardner, Silvio Savarese, Jeannette Bohg, and Animesh Garg.
\newblock Variable impedance control in end-effector space. an action space for reinforcement learning in contact rich tasks.
\newblock In {\em Proceedings of the International Conference of Intelligent Robots and Systems (IROS)}, 2019.

\end{thebibliography}


\clearpage
\newpage

\begin{figure*}[!t]
    \centering
    \includegraphics[width=0.8\linewidth]{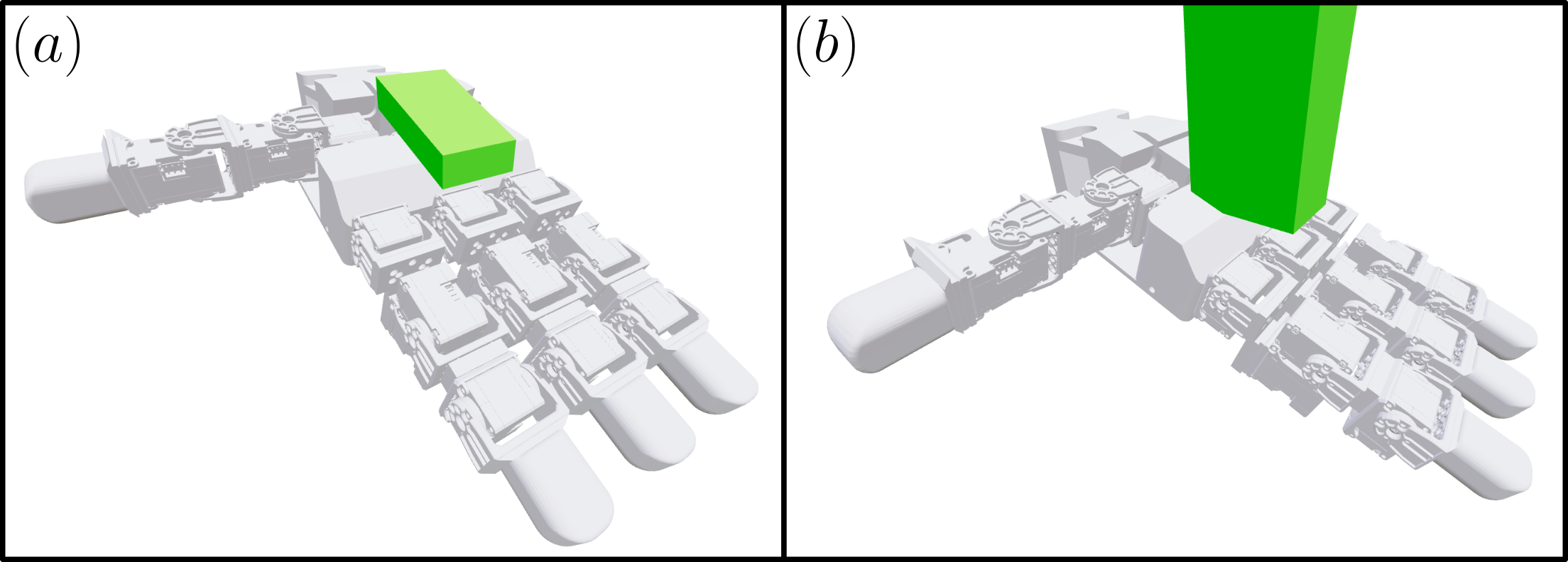}
    \caption{The safe regions $\mathcal{S}$ (green) for (a) when the cube's $xy$ coordinates lie in the palm, and (b) when the cube's $xy$ coordinates lie outside the palm.}
    \label{fig:safe_region}
\end{figure*}

\appendix

This appendix aims to make precise some of the technical details that were omitted from the main body of the paper for brevity. For the most fine-grained details, please see the open-source code provided at \href{https://caltech-amber.github.io/drop}{https://caltech-amber.github.io/drop}.

\subsection{Sampling-based Predictive Control Implementation}

\textbf{Warm-starts by time shifting.} Recall the generic SPC procedure described in Algorithm \ref{alg:sampling_planner}. Because the cost $J$ at some time depends on the current state $x_0=\hat{x}(t)$, once time elapses, the optimization problem \eqref{eqn:ocp} changes. Instead of solving the problem from scratch, the solution is \textit{warm started} by applying a time shift to the previous solution (e.g., as in \citeapp{sacks2023_dmpo}). For the last time step of the warm-started solution, we simply copy the second-to-last spline parameter.

To be precise, let $h$ denote an iteration of the mean control spline knots $\widebar{U}$. Then, applying the shift operation gives
\eqsnn{
    \widebar{U}[h] &= \brackets{u_1[h], u_2[h], \dots, u_K[h]}, \\
    \text{shift}\parens{\widebar{U}[h]} &= \brackets{u_2[h], u_3[h], \dots, u_K[h], u_K[h]},
}
which the new samples will be centered on (and similarly for any other parameters like sampling variances).

\textbf{Retaining the best rollout in PS.} In predictive sampling (PS), MJPC additionally keeps the current nominal (i.e., the best) control spline as one of the samples when resampling, so that if none of the sampled trajectories improves performance, the nominal policy is not updated. We found that this is critical to the performance of PS.

\textbf{Minimum standard deviation in CEM.} While the sampling variance is fixed in PS, in CEM, a diagonal covariance is repeatedly fit to the top $M$ elite samples. However, in practice, the variance may quickly collapse to very low values, preventing the planner from exploring. Therefore, we specify a minimum standard deviation uniformly over the diagonal entries of $\Sigma$, denoted $\sigma_\text{min}$.

\textbf{Default Parameters.} Table \ref{table:planner_parameters} lists the values of planner parameters used in experiments unless otherwise stated.

\begin{table}[h]
\centering
\begin{tabular}{|c|c|c|}
    \hline
    Method & Parameter & Value \\
    \hline
    \hline
    \multirow{4}{*}{All} & Spline Order & 0 \\
    & Plan Horizon & 1.0\unit{\s} \\
    & Plan Timestep & 0.01\unit{\s} \\
    & Num Spline Knots & 4 \\
    \hline
    \multirow{2}{*}{PS} & Num Samples & 120 \\
    & Sampling Stdev & 0.3 \\
    \hline
    \multirow{2}{*}{CEM} & Num Samples & 120 \\
    & Num Elites & 4 \\
    \hline
    iLQR & Num Parallel Line Search Threads & 120 \\
    \hline
\end{tabular}
\caption{Default planner parameter values used in all experiments.}
\label{table:planner_parameters}
\end{table}

\subsection{Specifics of the DROP Control Problem}

\textbf{The position cost term.} We now explain the exact form of the position cost term $\ell_\text{p}(p^c)$ in \eqref{eqn:pos_cost}. The function $\text{dist}_\mathcal{S}(\cdot)$ is parameterized by the ``safe region'' $\mathcal{S}$ in which we would like the cube's center to remain. Specifically, we have
\eqs{
    \text{dist}_\mathcal{S}(p^c) &:= d(\norm{p^c}_\mathcal{S}), \\
    d(x) &:= 0.05 \cdot \log\parens{1 + \exp\parens{\frac{250x}{0.05}}},
}
where $\norm{x}_\mathcal{S}$ reports the Euclidean distance between a point $x$ and a set $\mathcal{S}$, and $d(x)$ is a scalar-valued function that applies a very quickly-growing penalty when $x>0$. In particular, its parameters are chosen such that when $x=0.01$, $d(x) \approx 1$.

The set $\mathcal{S}$ is defined by two cases. If the cube's $x$ and $y$ coordinates lie within a specified rectangle, then we parameterize $\mathcal{S}$ as a parallelepiped whose $x$ and $y$ faces are axis-aligned, and whose $z$ faces are angled at the same angle as the palm of the hand, $\beta:=20^\circ$. Otherwise, we instead specify a uniform minimum $z$ height denoted $z^-$ and no maximum height. The second case is designed so that if the planner predicts the cube will leave the palm, it prioritizes not dropping it and returning it back to the palm. See Fig. \ref{fig:safe_region} for a visualization of these two cases.

The global origin of the system is located in the center of the interface between the hand mount and the bar of 80/20 to which it is affixed. The side length of the cube is $b:=0.07$. Let $(p^c_x, p^c_y, p^c_z)$ denote the position coordinates of the center of the cube. The dimensions associated with the figure are specified in Table \ref{table:safe_region}.
\begin{table}[h]
\centering
\begin{tabular}{|c|c|}
    \hline
    Dim & Value \\
    \hline
    $x^-$ & 0.08 \\
    $x^+$ & 0.14 \\
    $y^-$ & -0.02 \\
    $y^+$ & 0.02 \\
    $z^-$ & $\begin{cases}
        \frac{b}{2\cos(\beta)} - p^c_x \cdot \tan(\beta), & \text{if } p^c_x \in [x^-, x^+],\; p^c_y \in [y^-, y^+] \\
        -0.015, & \text{otherwise}
    \end{cases}$ \\
    $z^+$ & $\begin{cases}
        z^- + 0.035, & \text{if } p^c_x \in [x^-, x^+],\; p^c_y \in [y^-, y^+] \\
        +\infty, & \text{otherwise}
    \end{cases}$ \\
    \hline
\end{tabular}
\caption{Dimensions of the set $\mathcal{S}$.}
\label{table:safe_region}
\end{table}

\textbf{Cost Weights.} The weights in \eqref{eqn:ocp} are $\lambda_\text{g}=1.0, \lambda_\text{p}=2.5$.

\textbf{Time discretization.} While \eqref{eqn:generic_cost} is expressed as an integral over the interval $[0, T]$, in practice we numerically compute it by discretizing the interval using a fixed time step $\Delta t$:
\eqs{
    \int_0^T \ell\left(x(t), u(t)\right)dt &\approx \sum_{j=0}^H \ell\biggl(x(j\Delta t), u(j\Delta t)\biggr) \cdot \Delta t,
}
where $H=T / \Delta t$.

\subsection{Dataset and Training of the Keypoint Predictor}

\textbf{Additional dataset details.} We now further explain design decisions for the dataset used to train the vision model $g_\varphi$.

As described in Sec. \ref{sec:estimation}, the training images were generated per the Movi-F dataset convention from \href{https://github.com/google-research/kubric/tree/main/challenges/movi}{Kubric}. The images are sampled from frames of a video generated by simulating a random collection of objects (always including the cube) thrown into the scene from various positions and heights, with random camera motion and motion blur. The non-cube objects are drawn randomly from the \href{https://research.google/blog/scanned-objects-by-google-research-a-dataset-of-3d-scanned-common-household-items/}{Google Scanned Objects} dataset, and backgrounds are taken randomly from \href{https://polyhaven.com/hdris}{PolyHaven}. 

These data were split into train and test datasets by first separating videos into train and test videos then by randomizing the order of all images from these videos. This ensured that very similar frames from the same video did not appear in both the train and test sets. Each video supplied 24 images. In these videos, the cube's size was normalized such that its side length was 2.0\unit{\m}. Thus, when using the keypoint detector model downstream in conjunction with the smoother, the cube size must first be normalized.

Visual/material properties of the cube were randomized using \verb|kubric| by uniform random sampling (see Table \ref{table:kubric_augmentations}).
\begin{table}[h]
\centering
\begin{tabular}{|c|c|c|}
    \hline
    Property & Sampling Range \\
    \hline
    \hline
    Roughness & [0.0, 0.3] \\
    Specularity & [0.75, 1.0] \\
    Metallic & [0.25, 0.75] \\
    Index of Refraction & [1.0, 2.0] \\
    \hline
\end{tabular}
\caption{Cube material randomizations.}
\label{table:kubric_augmentations}
\end{table}

The camera pose was sampled such that it was always pointed at the origin, but its position was uniformly randomly located in some hemispherical shell with inner radius 7.0\unit{\m}, outer radius 9.0\unit{\m}, and $z$ height exceeding 0.2\unit{\m}.

As described in Sec. \ref{sec:estimation}, images in the dataset where too few or too many pixels corresponded to the cube were removed, as this indicated that the cube was either too occluded or too close to provide an accurate keypoint estimate. Specifically, an image was included in the dataset only if the proportion of cube pixels was in the interval $[0.02, 0.7]$.

Lastly, we note that while the size of the training dataset is fairly large for such a specific task (about 686000 images), it is far fewer than number of images used to train cube pose estimators in other works (e.g., 5 million in \cite{handa2024_dextreme} and we approximate 76.8 million in \cite{openai2018_dactyl}). However, it is clear that an improved different approach will need to be used in order to generalize to different objects.

\textbf{Image Augmentation Details.} One of the key factors for robust keypoint prediction in the real world was the application of extensive image augmentations during training. The augmentations were randomly sampled and applied during training using a mix of implementations from the open-source image processing package \verb|kornia| and custom augmentations described below. The \verb|kornia| augmentations and parameters are described in Table \ref{table:kornia_augmentations}. For detailed descriptions of the parameters, please see the \verb|kornia| documentation. Any time the parameter \verb|p| appears, it denotes the probability that the augmentation is applied.

\begin{table}[h]
\centering
\begin{tabular}{|c|l|}
    \hline
    \textbf{Augmentation} & \multicolumn{1}{c|}{\textbf{Parameters}} \\  
    \hline
    \hline
    \multirow{4}{*}{\texttt{RandomAffine}} & \texttt{degrees: 90} \\
     & \texttt{translate: (0.1, 0.1)} \\
     & \texttt{scale: (0.9, 1.5)} \\
     & \texttt{shear: 0.1} \\
    \hline
    \multirow{4}{*}{\texttt{RandomErasing1}} & \texttt{p: 0.5} \\
     & \texttt{scale: (0.02, 0.1)} \\
     & \texttt{ratio: (2.0, 3.0)} \\
     & \texttt{same\_on\_batch: False} \\
    \hline
    \multirow{4}{*}{\texttt{RandomErasing2}} & \texttt{p: 0.5} \\
     & \texttt{scale: (0.02, 0.05)} \\
     & \texttt{ratio: (0.8, 1.2)} \\
     & \texttt{same\_on\_batch: False} \\
    \hline
    \texttt{RandomPlanckianJitter} & \texttt{mode: blackbody} \\
    \hline
    \multirow{4}{*}{\texttt{ColorJiggle}} & \texttt{brightness: 0.2} \\
     & \texttt{contrast: 0.4} \\
     & \texttt{saturation: 0.4} \\
     & \texttt{hue: 0.025} \\
    \hline
    \multirow{3}{*}{\texttt{RandomGaussianBlur}} & \texttt{kernel\_size: (5, 5)} \\
     & \texttt{sigma: (3.0, 8.0)} \\
     & \texttt{p: 0.5} \\
    \hline
    \texttt{RandomPlasmaShadow} & All defaults \\
    \hline
\end{tabular}
\caption{All RGB image \texttt{kornia} augmentations used during training. Augmentation parameters are defaults if unspecified.}
\label{table:kornia_augmentations}
\end{table}

We implemented custom augmentations that applied only to depth images, as well as a custom augmentation for transplanting random backgrounds onto training images. We describe these now and summarize the associated parameters in Table \ref{table:custom_augmentations}.

For the depth readings, we (i) uniformly randomly sample a per-image bias (\verb|DepthBias|), (ii) add Gaussian noise (\verb|DepthGaussianNoise|), and (iii) uniformly randomly sample near and far planes centered about some mean value, where everything too close or far is set to 0.0 (\verb|DepthPlane|). 

Lastly, we implemented a special augmentation called \verb|CustomTransplantation| (not to be confused with the \verb|RandomTransplantation| augmentation in \verb|kornia|), which requires a batch of RGBD images as well as segmentation masks for the cube. With some probability, each image in the batch (the acceptor) has all of its non-cube pixels replaced by pixels from another image in the batch (the donor) using the following procedure. First, all non-cube pixels in the acceptor are identified, which initializes a background mask. Second, all pixels in the donor that are closer to the camera than the corresponding pixels in the acceptor are added to the background mask. Third, the pixels in the donor mask corresponding to the cube in the donor are removed from the mask (to prevent there from being two cubes in a single image). Fourth, the background mask selects pixels from the donor and replaces the corresponding pixels in the acceptor. Finally, if the resulting transplanted image results yields an image that has too large or small of a ratio of visible cube pixels, the transplantation is not applied. This transplantation method allows backgrounds from donors to further obscure the cube in acceptors in random ways, which helps improve robustness.

\begin{table}[h]
\centering
\begin{tabular}{|c|l|}
    \hline
    \textbf{Augmentation} & \multicolumn{1}{c|}{\textbf{Parameters}} \\  
    \hline
    \hline
    \multirow{2}{*}{\texttt{DepthBias}} & \texttt{bias\_range: (-0.02, 0.02)} \\
     & \texttt{p: 0.5} \\
    \hline
    \texttt{DepthGaussianNoise} & \texttt{stdev: 0.005} \\
    \hline
    \multirow{6}{*}{\texttt{DepthPlane}} & \texttt{near\_mean: 0.1} \\
     & \texttt{near\_range: (-0.05, 0.05)} \\
     & \texttt{p\_near: 0.5} \\
     & \texttt{far\_mean: 0.5} \\
     & \texttt{far\_range: (-0.05, 0.05)} \\
     & \texttt{p\_far: 0.5} \\
     \hline
    \multirow{3}{*}{\texttt{CustomTransplantation}} & \texttt{p: 0.5} \\
     & \texttt{ratio\_lb: 0.02} \\
     & \texttt{ratio\_ub: 0.7} \\
    \hline
\end{tabular}
\caption{Parameters for our custom augmentations.}
\label{table:custom_augmentations}
\end{table}

\textbf{Training Hyperparameters.} Training hyperparameters are described in Table \ref{table:training_hyperparameters}. We trained the model using a machine with 8 H100 GPUs for about 8 hours, though we remark that using this many GPUs is not required.

\begin{figure*}[t]
    \centering
    \includegraphics[width=\linewidth]{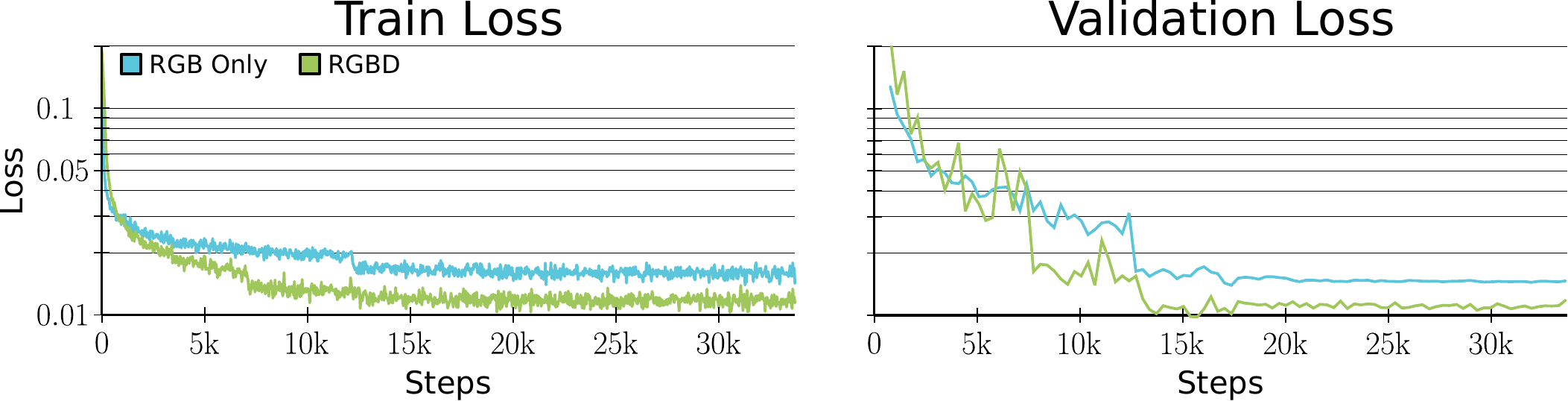}
    \caption{Training and validation loss curves for the RGBD and RGB-only vision models. Note that the curves are log scale.}
    \label{fig:loss_curves}
\end{figure*}

\begin{table}[h]
\centering
\begin{tabular}{|c|l|}
    \hline
    \textbf{Hyperparameter} & \multicolumn{1}{c|}{\textbf{Value}} \\
    \hline
    \hline
    \texttt{batch\_size} & 256 \\
    \hline
    \texttt{learning\_rate} (initial) & 1e-3 \\
    \hline
    \texttt{epochs} & 100 \\
    \hline
    \texttt{num\_workers} & 4 \\
    \hline
    \texttt{AMP} & \texttt{True} \\
    \hline
    \texttt{Loss Function} & \texttt{MSE} \\
    \hline
    \texttt{Optimizer} & \texttt{AdamW} \\
    \hline
    \multirow{5}{*}{\texttt{Scheduler}} & \texttt{ReduceLROnPlateau} \\
     & \texttt{patience: 5} \\
     & \texttt{factor: 0.25} \\
     & \texttt{min\_lr: 1e-6} \\
     & \texttt{grad\_scaler: True} \\
    \hline
\end{tabular}
\caption{Training Hyperparameters.}
\label{table:training_hyperparameters}
\end{table}

\textbf{Training and Validation Curves.} We show the training and validation curves for the RGBD model as well as the RGB-only model used in the ablation study from Sec. \ref{sec:hardware_ablations} in Fig. \ref{fig:loss_curves}. Overall, the RGBD model seems to perform slightly better, but not by a large margin. Note that the figure is in log scale.

\subsection{Smoother Details}\label{sec:smoother_impl}

This section provides a cursory overview of factor graphs, explains the setup of the graph used in DROP's estimation problem, and describes implementation details and hyperparameters of the smoother. For the most fine-grained explanation, please see our open-source code: \href{https://caltech-amber.github.io/drop}{https://caltech-amber.github.io/drop}.

\textbf{A primer on factor graphs.} Factor graphs are a type of probabilistic graphical model that models the relationships between unknown random variables via \textit{factors}, which can be interpreted as (unnormalized) likelihood functions of subsets of all of the unknowns parameterized by measurements, which themselves are functions of the same unknowns. The factor graph defines a factorization of some global likelihood function over all unknowns, which gives us a computationally-convenient framework for performing \textit{maximum a posteriori} (MAP) inference to recover an estimate of the unknowns. For further explanation, we refer the reader to \citeapp{dellaert2017_factorgraphs}.

Let $X$ denote the total set of all unknowns and $X_l$ denote some indexed subset of $X$. Then, we can represent the factors formally as
\eqs{
    \phi(X) = \prod_l \phi_l(X_l),
}
which models the independence relationships between the unknown variables of interest. Now, let measurements be modeled $z_l = h_l(X_l)$, where $h_l$ is some measurement model. Then, with a Gaussian noise model and the assumption that the factors take the form
\eqs{
    \phi_l(X_l) \propto \exp\braces{-\frac{1}{2}\norm{h_l(X_l) - z_l}_{\Sigma_l}^2},
}
the MAP solution for the unknown variables is equivalent to the following nonlinear least squares problem:
\eqs{\label{eqn:map_inference}
    X^\text{MAP} = \argmin_X\sum_l \norm{h_l(X_l) - z_l}_{\Sigma_l}^2.
}
In practice, the MAP inference problem is solved by numerical methods like the Gauss-Newton or Levenberg-Marquardt algorithms, which converge to some local minimum by solving successive linearizations of \eqref{eqn:map_inference}.

To reduce sensitivity to outlier measurements (i.e., keypoint detection failures), we used a \textit{robust Huber error model}  (see this GTSAM blog post for more information: \href{https://gtsam.org/2019/09/20/robust-noise-model.html}{gtsam.org/2019/09/20/robust-noise-model.html}).

\textbf{Incremental smoothing.} We are interested in the setting where we receive a stream of incremental information and we seek to update the MAP estimate over time. In particular, we consider a canonical graphical model of a dynamical system with two types of factors: \textit{dynamics} factors that model the (Markovian) temporal transitions between states ($\phi_t^\text{dyn}(X_{t+1}, X_t)$), and \textit{measurement} factors that model the likelihood of observations on those states ($\phi_t^\text{meas}(X_t;z_t)$).

Because the graph would rapidly grow in size as time elapses, when incrementally smoothing, it is common to \textit{marginalize} out the effect of older unknowns to retain their information content while removing them from the graph (see \citeapp[Sec. 5.3]{dellaert2017_factorgraphs}). This way, the number of variables remains fixed and the estimation problem stays tractable.

This scheme gives rise to \textit{fixed-lag smoothing}, where some lookback window specifies the number of states in the past to maintain in the graph at each time step. The case where the lookback is 1 corresponds to an iterated extended Kalman filter (iEKF), since marginalization happens after each measurement is received, and the MAP problem is solved through successive linearizations (see \citeapp[Sec 4.3.1]{barfoot2024state}).

\textbf{The cube pose estimation factor graph.} In DROP, we seek to estimate the cube poses (and velocities) over time. For computational speed, we do not model the full dynamics of the cube-hand system (which gives rise to the need for the corrector later). In particular, the stiff contact dynamics would introduce substantial numerical challenges when conducting successive linearizations to solve \eqref{eqn:map_inference}.

Instead, we use a \textit{constant velocity model} to approximate the cube dynamics. Because the linear and angular velocity of the cube is typically small when supported by the palm, this simple model (accompanied by a sufficiently tuned mistrust of it) provides a coarse but sufficiently-accurate model of the cube's motion.

For a given camera with a known global pose in conjunction with a pinhole camera model, the measurement model computes the expected keypoint locations in the camera's image frame (in pixel coordinates). The ``true'' measurements come from the keypoint predictor network $g_\varphi$.

\textbf{Sketch of the keypoint projection factor.} The keypoint factor (one for each keypoint, indexing suppressed for clarity) for a camera has the form
\eqs{
    \phi^\text{meas}_t(q^c_t; \nu^\text{cam}, p_\text{kp}^\text{pix}, p^c_\text{kp}, p^\text{cam}, r^\text{cam}),
}
where $q^c_t$ is the pose of the cube at time $t$, $\nu^\text{cam}$ is the camera's known intrinsics, $p_\text{kp}^\text{pix}\in\R^{2}$ is the measured keypoint locations in pixel coordinates from the predictor $g_\varphi$, $p^c_\text{kp} \in \R^3$ is the location of the associated keypoint in the cube's local frame, and $\parens{p^\text{cam}, r^\text{cam}} \in SE(3)$ is the known fixed pose of the camera in the world frame.

Let the function $\text{proj}: \R^3 \rightarrow \R^2$ project a spatial point expressed in the camera frame to a coordinate in pixel space, which is implemented in GTSAM \cite{dellaert2022_gtsam}. The measurement residual is then computed as follows:
\eqs{
    p_\text{kp}^\text{pix} - \text{proj}(p^c_\text{kp};\nu^\text{cam}, p^\text{cam}, r^\text{cam}),
}
where the Jacobians associated with frame transforms and the projection function are also computed by GTSAM. For the exact implementation details, please see our open-source implementation \cite{li2024_drop_website}.

\textbf{Smoother parameters and other details.} We summarize all parameters in Table \ref{table:smoother_params}. All quaternions are reported in wxyz order. Noise parameters are given by standard deviations $\sigma$ corresponding to diagonal components of the noise model covariance matrices. Rotational noise is represented by Gaussian noise in the tangent space. Huber regularization is applied identically to all noise models.
\begin{table}[h]
\centering
\begin{tabular}{|c|c|}
    \hline
    Parameter & Value \\
    \hline
    \hline
    \multirow{4}{*}{Cube Pos Prior} & $p=[0.1, 0.0, 0.0]$ \\
    & $r = [1.0, 0.0, 0.0, 0.0]$ \\
    & $\sigma_p = [0.5, 0.5, 0.5]$ \\
    & $\sigma_r = [1.0, 1.0, 1.0]$ \\
    \hline
    \multirow{4}{*}{Cube Vel Prior} & $v=[0.0, 0.0, 0.0]$ \\
    & $\omega=[0.0, 0.0, 0.0]$ \\
    & $\sigma_v = [0.01, 0.01, 0.01]$ \\
    & $\sigma_\omega = [0.2, 0.2, 0.2]$ \\
    \hline
    \multirow{2}{*}{Dynamics Pos Noise} & $\sigma_p = [0.01, 0.01, 0.01]$ \\
    & $\sigma_r = [0.2, 0.2, 0.2]$ \\
    \hline
    \multirow{2}{*}{Dynamics Vel Noise} & $\sigma_v = [0.01, 0.01, 0.01]$ \\
    & $\sigma_\omega = [0.2, 0.2, 0.2]$ \\
    \hline
    Measurement Noise (in pixels) & $\sigma_z = [3.0, 3.0]$ \\
    \hline
    Meas. Robustness Parameter $k_\text{huber}$ & 1.345 \\
    \hline
    Lookback Window & 1 \\
    \hline
\end{tabular}
\caption{Smoother parameters.}
\label{table:smoother_params}
\end{table}

\subsection{Corrector Implementation}
As describe in Sec. \ref{sec:estimation}, the corrector applies a virtual corrective wrench to its internal estimate of the cube state as shown in \eqref{eqn:corrector_wrench}. The corrector gain matrices have 6 diagonal entries corresponding to the virtual forces and torques applied to the cube. We use the values
\eqs{
    C_P &= \diag(\brackets{1000, 1000, 1000, 3, 3, 3}), \\
    C_D &= \diag(\brackets{1, 1, 1, 0.001, 0.001, 0.001}).
}
The extra gravitational corrective force described in Sec. \ref{sec:estimation} is simply an extra 10\unit{\N} added along the $-z$ axis of the virtual forces applied to the cube.

Additionally, there are a few parameters of the corrector dynamics $\hat{f}$ that differ from the planner's internal model $f$, which we show in Table \ref{table:corrector_model_params}.

\begin{table}[h]
\centering
\begin{tabular}{|c|c|}
    \hline
    Parameter & Value \\
    \hline
    \hline
    Timestep & 0.04\unit{\s} \\
    LEAP Hand $k_p$ & 3.0 \\
    \texttt{solimp} & [0.999, 0.999, 0.001, 0.0001, 1] \\
    \texttt{solref} & [0.0001, 1] \\
    \hline
\end{tabular}
\caption{Parameters of the corrector's internal model $\hat{f}$ that differ from the planner's internal model $f$.}
\label{table:corrector_model_params}
\end{table}

\subsection{Implementation of Noise Random Walk}
In Sec. \ref{sec:simulation}, we explained that when corrupting the perception stack, we simulate additive noise using a bounded random walk. The reason for this choice is that in real life, perception errors are highly-correlated with state. For example, if there is a 1\unit{\cm} error in cube keypoint estimates at some cube pose, then we typically observe that the error would remain at similar levels with low noise so long as the cube does not move much from that state. Thus, simply adding 0-mean uncorrelated noise with high variance is not an accurate simulated representation of what we observe in the real world.

Instead, by modeling the noise as a bounded random walk with very low noise, we can approximately model the effects of sustained pose estimation bias. Specifically, let $\nu \in \R^3, \xi \in \mathfrak{so}(3) \cong \R^3$ denote the position and rotational noises respectively. These noise values are iteratively updated as follows, where $\Delta t$ represents whatever amount of time has elapsed since the last update:
\eqs{
    \nu(t+\Delta t) &= \text{clip}\parens{\nu(t) + z_\nu(t), \nu_\text{lb}, \nu_\text{ub}}, \\
    \xi(t+\Delta t) &= \text{clip}\parens{\xi(t) + z_\xi(t), \xi_\text{lb}, \xi_\text{ub}},
}
where $\nu_\text{lb}, \nu_\text{ub}, \xi_\text{lb}, \xi_\text{ub}$ are fixed lower and upper bounds for the random walk, $\text{clip}(\cdot)$ clips the values in the first argument elementwise to the provided lower and upper bounds, and $z_nu\sim\mathcal{N}(0, \Sigma_\nu), z_xi\sim\mathcal{N}(0, \Sigma_\xi)$ are randomly-sampled noises that drive the random walk.

Finally, we update the cube pose $q^c = [p^c, r^c] \in SE(3)$ as follows:
\eqs{
    p^c(t + \Delta t) &= p^c(t) + \nu(t + \Delta t) \\
    r^c(t + \Delta t) &= r^c(t) \circ \text{Exp}\parens{\xi(t+\Delta t)}.
}
The parameters we used are summarized in Table \ref{table:random_walk}.

\begin{table}[h]
\centering
\begin{tabular}{|c|c|}
    \hline
    Parameter & Value \\
    \hline
    \hline
    $\nu_\text{lb}$ & [-0.01, -0.01, -0.01] \\
    $\nu_\text{ub}$ & [0.01, 0.01, 0.01] \\
    $\xi_\text{lb}$ & [-0.1, -0.1, -0.1] \\
    $\xi_\text{ub}$ & [0.1, 0.1, 0.1] \\
    \hline
    $\Sigma_\nu$ & \text{diag}([0.001, 0.001, 0.001]) \\
    $\Sigma_\xi$ & \text{diag}([0.0001, 0.0001, 0.0001]) \\
    \hline
\end{tabular}
\caption{Parameters of the noise random walk.}
\label{table:random_walk}
\end{table}

\subsection{Miscellaneous Hardware Implementation Details}
While the stack described in Sec. \ref{sec:method} is fairly simple, there are a few miscellaneous details that affect performance.

\textbf{LEAP hand gains.} The scale of the gains in the simulated model and on hardware are quite different. While in simulation, the planner's LEAP hand gains are $k_p=1.0$ and $k_v=0.01$, on hardware, we use $k_p=75$ and $k_v=25$. In fact, the hardware gain values are already quite low compared to the values recommended in \cite{shaw2023_leap}. This was intentional; by lowering the gains, we lowered the amount of energy imparted on the cube, and allowed the controller to manipulate the cube with a higher degree of control.

\textbf{The physical cube.} The cube itself is made of low-density 3D-printed PLA, weighing 0.108\unit{\kg}. The faces of the cube are simply printed out on regular printer paper and taped onto the cube using reflective, low-friction packing tape. We remark that over time, the frictional properties of the cube may slightly change. Our reported experiments were run using a cube that was used for about 1 week prior, and anecdotally, the friction increased slightly over this period, which helped prevent the cube from slipping out of the hand. In the weeks prior to our final experimental trials, older versions of the cube also exhibited peeling tape, torn faces/edges, and other effects that substantially increased the friction of the cube. For the most repeatable results, we would recommend re-taping a cube from scratch every few weeks.

\textbf{Manual calibration of camera poses.} Before running experiments, we found it important to perform some manual adjustments of each camera's ground-truth pose with respect to the world frame. To perform the calibration, we streamed the estimated cube pose from the smoother and projected the associated analytical keypoint locations back onto each camera's image frame using the keypoint factors derived in App. \ref{sec:smoother_impl}. By adjusting the camera position, we were able to manually align the projected keypoints with the corners of the cube.

\textbf{Manual adjustments to the smoothed cube pose.} We found that errors in the camera poses manifested as nearly-constant translational biases in the smoothed cube pose estimate. To remedy this, we simply manually adjusted estimated cube pose by adding an appropriate offset.

\subsection{Acknowledgments}
We thank Taylor Howell, Tom Erez, and Yuval Tassa for early discussions regarding sampling-based trajectory optimization for contact-rich tasks. We thank Taylor Howell for continual support on MJPC features that were key for the development of our work.

We thank Gavin Hua for helping to write low-level code for interfacing with the LEAP hand. We thank Zach Olkin for general assistance with setting up the control stack and implementation advice.

We thank Simon Le Cleac'h for insightful discussions and advice on contact-aware state estimation and the corrector.

We thank Sabera Talukder for assisting with the construction of cubes used in our hardware trials.

Lastly, we thank the creators and maintainers of all open source software that made this paper possible, including the following, which were not cited in the main body: \verb|pytorch| \citeapp{paszke2019_pytorch} and \verb|pybullet| \citeapp{coumans2021_pybullet}.

\bibliographystyleapp{unsrt}
\bibliographyapp{references}

\end{document}